\documentclass[10pt,twocolumn,letterpaper]{article}

\usepackage{cvpr}              

\usepackage{graphicx}
\usepackage{amsmath}
\usepackage{amssymb}
\usepackage{booktabs}
\usepackage{float}
\usepackage{makecell}
\usepackage{mathtools}
\usepackage{ragged2e}
\usepackage{pifont}
\usepackage{soul,color}

\usepackage[pagebackref,breaklinks,colorlinks]{hyperref}
\usepackage{graphicx}

\usepackage[capitalize]{cleveref}
\crefname{section}{Sec.}{Secs.}
\Crefname{section}{Section}{Sections}
\Crefname{table}{Table}{Tables}
\crefname{table}{Tab.}{Tabs.}

\def\cvprPaperID{5579} 
\def\confName{CVPR}
\def\confYear{2023}

\begin{document}

\title{Separated RoadTopoFormer}

\author{Mingjie Lu\footnotemark[1], Yuanxian Huang\footnotemark[1], Ji Liu, Jinzhang Peng, Lu Tian, Ashish Sirasao\\
Advanced Micro Devices, Inc., Beijing, China\\
{\tt\small (Mingjie.Lu, YuanXian.Huang, Ji.Liu, jinz.peng, lu.tian, ashish.sirasao)@amd.com}
}
\maketitle
\renewcommand{\thefootnote}{\fnsymbol{footnote}}
\footnotetext[1]{These authors contributed equally to this work.}

\begin{abstract}
Understanding driving scenarios is crucial to realizing autonomous driving. Previous works such as map learning and BEV lane detection neglect the connection relationship between lane instances, and traffic elements detection tasks usually neglect the relationship with lane lines. To address these issues, the task is presented which includes 4 sub-tasks, the detection of traffic elements, the detection of lane centerlines, reasoning connection relationships among lanes, and reasoning assignment relationships between lanes and traffic elements. We present Separated RoadTopoFormer to tackle the issues, which is an end-to-end framework that detects lane centerline and traffic elements with reasoning relationships among them. We optimize each module separately to prevent interaction with each other and aggregate them together with few finetunes. For two detection heads, we adopted a DETR-like architecture to detect objects, and for the relationship head, we concat two instance features from front detectors and feed them to the classifier to obtain relationship probability. Our final submission achieves 0.445 OLS, which is competitive in both sub-task and combined scores.
\end{abstract}

\section{Introduction}

In recent years, the availability of public large-scale datasets and benchmarks has greatly facilitated autonomous driving research. Many datasets \cite{chen2022persformer, yu2020bdd100k} focus on sensing visible lane lines to keep vehicles on the right track only, or to obtain traffic information by detecting traffic signals only.
However, the separation of tasks leads to a limited understanding of driving scenarios.
For example, a driving vehicle will be confused when it sees a green light but the lane it follows is controlled by another red light.
Based on this limitation, a key aspect of this task \cite{wang2023openlanev2} is to understand the complex driving environment, which is a prerequisite for making reasonable decisions.
On the one hand, this task wants to establish a strong association between traffic elements and lanes. On the other hand, understanding the separations between neighboring lanes is also necessary for guiding the vehicle driving on the desired trajectory. Both topology reasoning tasks are extremely challenging.

This task can be divided into two parts simply,  which are scene structure perception and reasoning. The scene structure perception aims to find out what and where the traffic elements and lanes are and the reasoning aims to understand the relationship between them. The latter is highly dependent on the former, but the reverse is not certain. So, we optimize each module separately to prevent interactions during training, and finally integrate them by finetuning.
Experiments prove it works. We also have made other experimental improvements, please refer to Section \ref{Methods}.

\begin{figure*}[!ht]
\begin{center}
\includegraphics[width=0.9\textwidth]{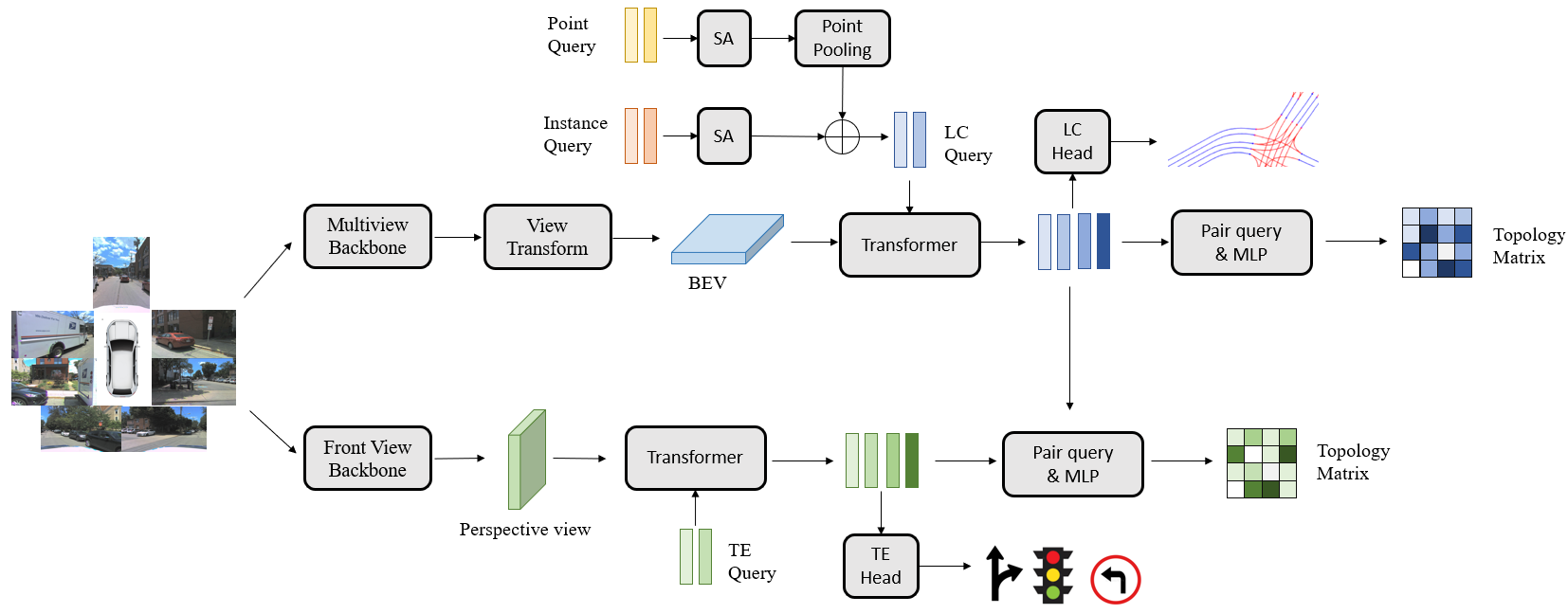}
\end{center}
   \caption{Overall framework.}
\label{fig:framework}
\end{figure*}

\section{Datasets}
Road Genome, also known as OpenLane-V2 \cite{wang2023openlanev2}, is the first dataset focusing on topology reasoning in the autonomous driving area. It contains 2.1M instance-level annotations and 1.9M positive topology relationships. This challenge is based on subset\_A, which contains 22477 training frames, 4806 val frames, and 4816 test frames. Each frame contains 6 surrounding images with resolution $1550 \times 2048$ and a front-view image with resolution $2048 \times 1550$. The final metric is OpenLane-V2 Score (OLS), which is the average of various metrics from different subtasks and is defined to describe the overall performance of the primary task: $OLS = \frac{1}{4}[DET_{l} + DET_{t} + f(TOP_{ll}) + f(TOP_{lt})]$, where $f$ is a scaling function that balances the scale of different metrics.

\section{Methods} \label{Methods}
\subsection{Baseline}
The official baseline \cite{wang2023openlanev2} provides a simple and easy-to-follow framework that
generates two feature maps from different views. One is in BEV (Bird's-eye view) and the other one is in PV (Perspective view). The former is used to predict lane centerlines (LCs) and the latter is for traffic elements (TEs) prediction. Two detection heads adopt similar DERT-like architectures. The following two relationship prediction modules  establish pairwise relationships, which contain a $L \times L$ lanes relationship matrix and a $L \times T$ lanes-traffic elements relationship matrix, where $L$ and $T$ represent the numbers of LCs and TEs from front detected results . Then two subsequent MLPs are used to predict the logits of two kinds of relationships, respectively. 

\begin{table}
\begin{center}
\begin{tabular}{|c|c|}
\hline
Method & DET$_l$ ( \% ) \\
\hline\hline
Baseline & 9.57 \\
+ decoupled training & 10.59 \\
+ 11 points representation & 13.58 \\
+ swin-s \& rescale & 22.19	\\
+ finetune with smaller lr & 23.44 \\
+ hierarchical query & 23.80 \\
+ intersection-sensitive & 25.87 \\
+ finetune whole model & 26.95 \\
\hline
\end{tabular}
\end{center}
\caption{Ablation of 3D centerline detection on OpenLane-V2 validation set.}
\label{tab:lc_performance}
\end{table}

\begin{table}
\begin{center}
\begin{tabular}{|c|c|}
\hline
Method & DET$_t$ ( \% ) \\
\hline\hline
Baseline & 45.89 \\
+ swin-s \& decoupled training & 58.41	\\
+ DINO head & 60.23 \\
+ finetune whole model & 61.42 \\
\hline
\end{tabular}
\end{center}
\caption{Ablation of traffic element detection on OpenLane-V2 validation set.}
\label{tab:te_performance}
\end{table}

\begin{table}
\begin{center}
\begin{tabular}{|c|c|}
\hline
Method & TOP$_{ll}$ ( \% ) \\
\hline\hline
Baseline & 0.92 \\
+ better LC detector & 3.90 \\
+ geometric clues & 14.26 \\
+ finetune whole model & 15.37 \\
\hline
\end{tabular}
\end{center}
\caption{Ablation of topology prediction between lane centerlines on OpenLane-v2 validation set.}
\label{tab:lclc}
\end{table}


\begin{table*}[!h]
\begin{center}
\begin{tabular}{|c|c|c|c|c|c|}
\hline
Method & DET$_l$ ( \% ) & DET$_t$ ( \% ) & TOP$_{ll}$ ( \% ) & TOP$_{lt}$ ( \% ) & OLS ( \% )\\
\hline\hline
Baseline \cite{wang2023openlanev2} & 9.57 & 45.89 & 0.92 & 11.46 & 24.72 \\
Ours & 26.95 & 61.42 & 15.37 & 21.81 & 43.57 \\
\hline
\end{tabular}
\end{center}
\caption{Submission results on OpenLane-V2 validation set.}
\label{tab:final_performance_val}
\end{table*}

\begin{table*}[hbt]
\begin{center}
\begin{tabular}{|c|c|c|c|c|c|}
\hline
Method & DET$_l$ ( \% ) & DET$_t$ ( \% ) & TOP$_{ll}$ ( \% ) & TOP$_{lt}$ ( \% ) & OLS ( \% )\\
\hline\hline
TopoNet \cite{li2023toponet} & 19 & 58 & 2 & 16 & 33 \\
Ours & 22 & 72 & 13 & 23 & 45 \\
\hline
\end{tabular}
\end{center}
\caption{Submission results on OpenLane-V2 test set.}
\label{tab:final_performance_test}
\end{table*}

\subsection{Architecture}

The design of our algorithm follows Road Genome \cite{wang2023openlanev2}. However, unlike Road Genome, our TE branch and LC branch do not share a common backbone as demonstrated in Figure \ref{fig:framework}. Instead, each branch has an independent backbone network to extract features. This modification allows for independent feature learning and data augmentation for two detection tasks. 

\textbf{Lane centerline detection}. Given multi-view images, we first use a shared Swin-small \cite{liu2021Swin} backbone to extract features from each view's image. Then, we apply BEVFormer\cite{li2022bevformer} to transform the multi-perspective view features into a unified BEV feature. Later, a Deformable DETR-like\cite{zhu2020deformable} transformer is utilized to extract query-wise information of the 3D lane centerlines based on the BEV feature. Finally, each output query is passed through an LC head to predict the confidence of a line and the coordination of 11 equally spaced 3D points in the centerline. The coordination of each 3D point is normalized according to the detection range.


\textbf{Traffic element detection}. We utilize a separated and independent Swin-small backbone to extract the perspective view feature from the front center image. DINO\cite{zhang2023dino} head is employed to detect 2D traffic elements.

\textbf{Topology prediction}. We follow the design of topology prediction in STSU \cite{Can2021StructuredBT}. Every two objects' query will be concatenated. The concatenated feature will pass through an MLP and a sigmoid layer and output a relationship confidence. The two objects will be considered as having a topology relationship only if the confidence is greater than 0.5. Instead of considering all queries like the baseline, we only consider the query whose confidence is bigger than a prior threshold.

\subsection{Bells and whistles}

\textbf{Hierarchical query}. For 3D centerlines detection, the locations of points are significant for the final performance. We design two kinds of queries, point query and instance query, to make the query input transformer decoder have better representation ability. Point queries $Q_p \in \mathbb{R}^{N_p \times D}$ and instance queries $Q_I \in \mathbb{R}^{N \times D}$ are first passed through a self-attention module to model the relationship between queries, where $N_p$ represents the number of point queries and is set to 11 to be equal to the final output number of points, $N$ represents the max number of centerlines, and $D$ represents the dimension of the embeddings. To aggregate the feature of both kinds of queries, a point pooling module is proposed to get a global feature across point queries. We utilize the sum operation to pool the point queries. Finally, LC query $Q_{LC}$ is obtained by adding each instance query to the 3D global pooling point feature.

\begin{equation}
Q_{pooled} = PointPooling(Q_p) = \sum\limits_{i=1}^{N_p}{Q_{p,i}}, Q_{p,i} \in \mathbb{R}^{D}
\end{equation}

\begin{equation}
Q_{LC,i} = Q_{I,i} + Q_{pooled}
\end{equation}

\textbf{Intersection-sensitive classification head}. The OpenLane-V2 \cite{wang2023openlanev2} dataset contains two kinds of centerline, normal lane centerline and connecting line in intersections, which are evidently different. Unlike normal lane centerlines with obvious 
local texture features, connecting lines in the intersection are virtual lines, which are used to describe the relationship among normal lane centerlines. Therefore, we distinguish these two categories in the classification head in the LC head. As shown in Table \ref{tab:lc_performance}, this simple strategy improves the DET\_l metric by 2.43\%.

\textbf{Swin backbone and input resolution}. Because the input image size of the baseline \cite{wang2023openlanev2} is the original resolution of the image, which is 1550x2048, the batch size can only be set to one on every GPU when training the whole model. However, the backbone of the baseline is ResNet50 \cite{He2015DeepRL} and utilizes BatchNorm, which is inappropriate when the batch size is set to one. Therefore, we utilize Swin-small \cite{liu2021Swin} as our backbone for both LC branch and TE branch, which apply LayerNorm instead of BatchNorm. Besides, to speed up the training and save device memory, we resize the multiview images to 775x1024. For the front view image, we keep its size as its original resolution (2048x1550), because its overhead is affordable. The backbones in both two branches are pre-trained in ImageNet1K \cite{imagenet}.

\textbf{11 points representation}. Instead of representing the 3D line as five Bezier control points like STSU \cite{Can2021StructuredBT}, we directly model the 3D line as 11 equally spaced keypoints in its skeleton. We found this simple representation is surprisingly better than the Bezier curve. Results are shown in Table \ref{tab:lc_performance}. 

\textbf{DINO TE detector}. We use the DINO\cite{zhang2023dino} detector head instead of the original deformable-detr of the baseline with 900 queries. As show in Table \ref{tab:te_performance}, DINO brings about a 2\% gain for traffic elements detection. 

\textbf{Geometric clues for relationship prediction between centerlines}. The topological relationship between centerlines is not only related to semantic information but also associated with their geometric locations. If the endpoints of the centerlines of two lanes are very close, then there is a high probability that they are topologically related. Therefore, we introduce geometric clues for relationship prediction between centerlines in two aspects. First, we concatenate the LC query with its start point and end point which are predicted by the LC regression head. Second, any two lane centerlines whose start and end points are less than three meters apart will be considered to have a topological relationship, even if their relationship confidence is less than 0.5. Results are shown in Table \ref{tab:lclc}.

\textbf{Decoupled training and integrated finetuning}. Instead of training all modules of the whole network simultaneously, we decouple different modules and train only one of them each time. Specifically, we first independently train the LC module and TE module. Then, two relationship heads are trained with frozen backbones and detection heads. The decoupled training strategy helps us quickly verify an improvement idea for a single module. Meanwhile, this strategy enables each module to perform its own duties and avoids the impact between different tasks. After all modules are trained independently, we finetune the whole network with a smaller learning rate. During finetuning, only four heads are unfrozen, including the LC head, TE head, and two relationship heads. In the decoupled training set, we follow the training setting in Road Genome \cite{wang2023openlanev2}, including the optimizer, the learning rate update schedule, and so on. The learning rate will be adjusted proportionally with the batch size. In the finetuning stage, we set a smaller learning rate, which is a quarter of the decoupled training stage.

\section{Final Results}

For the final submission, we apply all the aforementioned strategies for performance improvement. The performances on the OpenLane-V2 validation and test set are demonstrated in Table \ref{tab:final_performance_val} and  \ref{tab:final_performance_test}, respectively.

{\small
\bibliographystyle{ieee_fullname}
\bibliography{egbib}
}
\end{document}